\documentclass{egpubl}
\usepackage{eurovis2021}

\SpecialIssuePaper         

\usepackage[T1]{fontenc}
\usepackage{dfadobe}  

\usepackage{cite}  
\BibtexOrBiblatex
\electronicVersion
\PrintedOrElectronic
\ifpdf \usepackage[pdftex]{graphicx} \pdfcompresslevel=9
\else \usepackage[dvips]{graphicx} \fi

\usepackage{egweblnk}

\usepackage[ruled]{algorithm2e}
\usepackage{algpseudocode}  
\usepackage{amsmath}  

\usepackage{multirow}
\usepackage{diagbox}

\usepackage{url}

\usepackage{soul}

\title{Exploring Multi-dimensional Data via Subset Embedding}

\author[P.Xie, W. Tao, J. Li, W. Huang \& S. Chen]
{\parbox{\textwidth}{\centering Peng Xie$^{1}$\orcid{0000-0003-4270-8122}, Wenyuan Tao$^{1}$\orcid{0000-0001-9047-4231}, Jie Li$^{1}$\thanks{Jie Li is the corresponding author. Email: jie.li@tju.edu.cn}\orcid{0000-0001-6511-4090 }, Wentao Huang$^{1}$\orcid{0000-0003-2976-7953}
        and Siming Chen$^{2}$\orcid{0000-0002-2690-3588}}
        \\
{\parbox{\textwidth}{\centering $^1$College of Intelligence and Computing, Tianjin University, Tianjin, China\\
         $^2$School of Data Science, Fudan University, Shanghai, China
       }
}
}

\begin{document}


\maketitle
\begin{abstract}
Multi-dimensional data exploration is a classic research topic in visualization. Most existing approaches are designed for identifying record patterns in dimensional space or subspace. In this paper, we propose a visual analytics approach to exploring subset patterns. The core of the approach is a subset embedding network (SEN) that represents a group of subsets as uniformly-formatted embeddings. We implement the SEN as multiple subnets with separate loss functions. The design enables to handle arbitrary subsets and capture the similarity of subsets on single features, thus achieving accurate pattern exploration, which in most cases is searching for subsets having similar values on few features. Moreover, each subnet is a fully-connected neural network with one hidden layer. The simple structure brings high training efficiency. We integrate the SEN into a visualization system that achieves a 3-step workflow. Specifically, analysts (1) partition the given dataset into subsets, (2) select portions in a projected latent space created using the SEN, and (3) determine the existence of patterns within selected subsets. Generally, the system combines visualizations, interactions, automatic methods, and quantitative measures to balance the exploration flexibility and operation efficiency, and improve the interpretability and faithfulness of the identified patterns. Case studies and quantitative experiments on multiple open datasets demonstrate the general applicability and effectiveness of our approach.


\end{abstract}  



\section{Introduction}

Multi-dimensional data exploration is a classic research topic in visualization. Most existing approaches work at the record level, using various machine learning algorithms to find distinctive distributions (e.g. clusters or outliers) in dimensional space or subspace as data patterns \cite{yuan2013dimension, pagliosa2015projection, xia2017ldsscanner}. 

\begin{figure}[tb]
 \centering
 \includegraphics[width=\columnwidth]{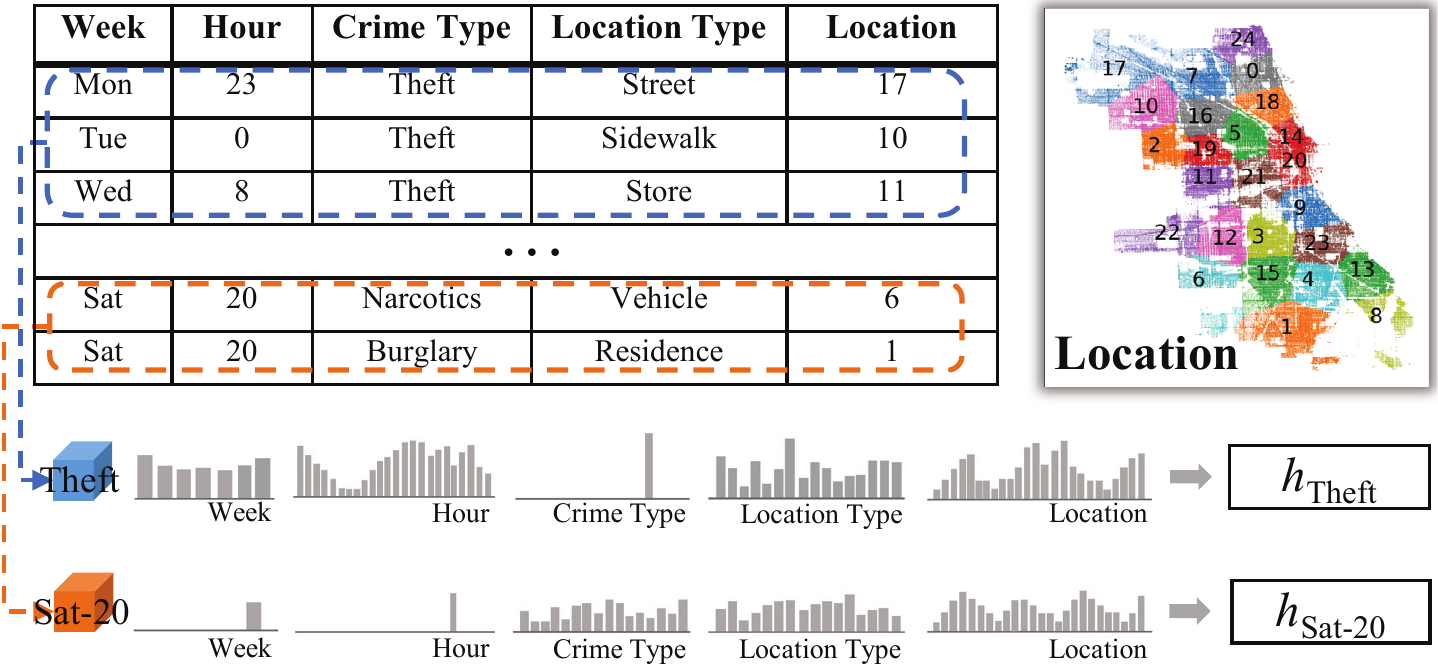}

 \caption{Two subsets of the Chicago crime dataset \cite{chicagocrime}. Each consists of records (crimes) with specified attribute values, and takes the distribution of the number of records on an attribute as a feature. The purpose of SEN is representing a large number of subset as uniformly-formatted vectors.}
 \label{fig:pic1}
\end{figure}

Patterns of multi-dimensional data, however, are often related to subsets rather than records. A \textit{\textbf{subset}} consists of data records and has multiple features, as in Figure \ref{fig:pic1}. Each \textit{\textbf{feature}} reflects an aspect of statistical information of all the included records. Typically, a \textit{\textbf{subset pattern}} is a group of subsets having similar values on specific features. We consider exploring subset patterns a more general task for multi-dimensional data, since a subset can only include one data record in an extreme setting. In that case, the subset pattern exploration degenerates to the record pattern exploration. 

There are many ways to partition a given multi-dimensional dataset into subsets and a subset can have a large number of features. Subset patterns, however, may associate with a few subsets and features. Without prior knowledge, analysts have to attempt different partition methods, repeatedly select a portion of generated subsets, and correlate them on different combinations of features to identify patterns. The huge search space makes the discovery of subset patterns a challenging and time-consuming process. 

The great success of representation learning \cite{bengio2013representation} inspires us to apply a subset embedding network (SEN) in multi-dimensional data exploration. The SEN can generate uniformly-formatted embeddings for a group of given subsets, as in Figure \ref{fig:pic1}. Moreover, the embedding similarity reflects that of subsets in terms of their feature values. We thus can perform automatic algorithms, such as clustering and dimensionality reduction (DR), according to their embeddings to identify patterns in an efficient manner.

In this paper, we propose the SEN. The design challenges include 1) adapting the SEN to arbitrary subsets to achieve better applicability (R1), 2) capturing the similarity on single features to accurately encode patterns (R2), and 3) obtaining a high training efficiency to enable incorporation into visualization systems (R3). The three aspects prevent the application of existing techniques in subset embedding (Section \ref{sec:section2.2}). Inspired by multi-view learning \cite{li2018survey}, we consider a subset a multi-view object and each subset feature a snapshot taken by a virtual camera around the subset (Section \ref{sec:section2.3}), and propose SEN of multi-subnet structure to satisfy the three requirements.

We integrate the SEN into a visualization system to explore subset patterns in multi-dimensional data. The system follows an “overview->details->patterns” explorative workflow \cite{shneiderman1996eyes}: analysts slice the data into subsets, project subsets according to their embeddings obtained from a SEN trained on-the-fly, and select portions of subsets to determine the existence of patterns. The system contains three components. Specifically, one utilizes the tree metaphor to achieve progressive data partition, which enables analysts to generate a variety of subsets through few operations. Another incorporates interactions and automatic methods to assist in exploring the projection. The third one provides a group of views implemented as classic visualization techniques for visualizing features of selected subsets to identify patterns. Generally, the three components balance the exploration flexibility and operation efficiency, and improve the interpretability and faithfulness of the identified patterns.

We conduct case studies and quantitative experiments to evaluate the approach on six open datasets. Experiment results illustrate its general applicability and effectiveness. Specifically, analysts can identify rich patterns using our approach by conducting tasks on drastically different subsets. Meanwhile, quantitative experiments demonstrate the high training efficiency and the effectiveness in capturing patterns of the SEN.

The main contribution of our work is a visualization approach to exploring subset patterns in multi-dimensional data, which integrates 1) \textit{\textbf{a subset embedding network}} that can accurately and quickly represent a large number of given subsets as uniformly-formatted embeddings , and 2) \textit{\textbf{a visualization system}} following a classic workflow, which combines 1) with three visual components to implement flexible and efficient subset pattern exploration.

The rest of the paper is organized as follows. Section \ref{sec:section2} gives design requirements. Sections \ref{sec:section3} and \ref{sec:section4} introduce the embedding network and the visualization system. Sections \ref{sec:section5} demonstrates the usability of our approach through case studies and quantitative experiments. Section \ref{sec:section6} discusses the limitations. We reviews the related work in Section \ref{sec:section7} and conclude the paper in Section \ref{sec:section8}.
\section{Problem Statement}
\label{sec:section2} 
 We give the subset definition, identify three  requirements, and outline the approach.

\subsection{Subset Definition}
Let $D(d_{1}, …, d_{n})$ be a multi-dimensional dataset, where $d_{i}$ is an attribute with the domain $dom(d_{i})$. A \textit{\textbf{subset}} consists of records selected by a group of filters, i.e. {$r(d_{1})$,…,$r(d_{n})$}, where $r(d_{i})$ is a filter that specifies a value range on $dom(d_{i})$, i.e. $r(d_{i})\in dom(d_{i})$.

Filters whose value ranges cover the whole domain, i.e. $r(d_{i})=dom(d_{i})$, can be hidden for brevity. We call the attribute of an unhidden filter a \textit{\textbf{slicing attribute}}. For example, in Figure \ref{fig:pic1}, the subset (Crime-type: theft) has a single slicing attribute, i.e. Crime-type, while the subset (Week: Saturday, Hour: 20) takes Week and Hour as two slicing attributes.

The number of unhidden filters describes the \textit{\textbf{dimensionality}} of the subset, denoted as $l=\left | \left \{ r(d_{i})|r(d_{i}) \ne dom(d_{i}) \right \}  \right | $. For example, (Crime-type: theft) is a 1-dimensional subset, and (Week: Saturday, Hour: 20) is a 2-dimensional subset.

Each subset can have multiple \textit{\textbf{features}}, as in Figure \ref{fig:pic1}. We write the $v$th feature of subset $S_{i}$ as $X_{i}^{(v)}$. A feature, describing an aspect of statistical information of all the included records, can be in any form, such as a number (e.g. count of all included records) or a vector (e.g. distribution of aggregate values on an attribute). 

\subsection{Design Requirements}
\label{sec:section2.2} 
\begin{figure}[tb]
 \centering
 \includegraphics[width=\columnwidth]{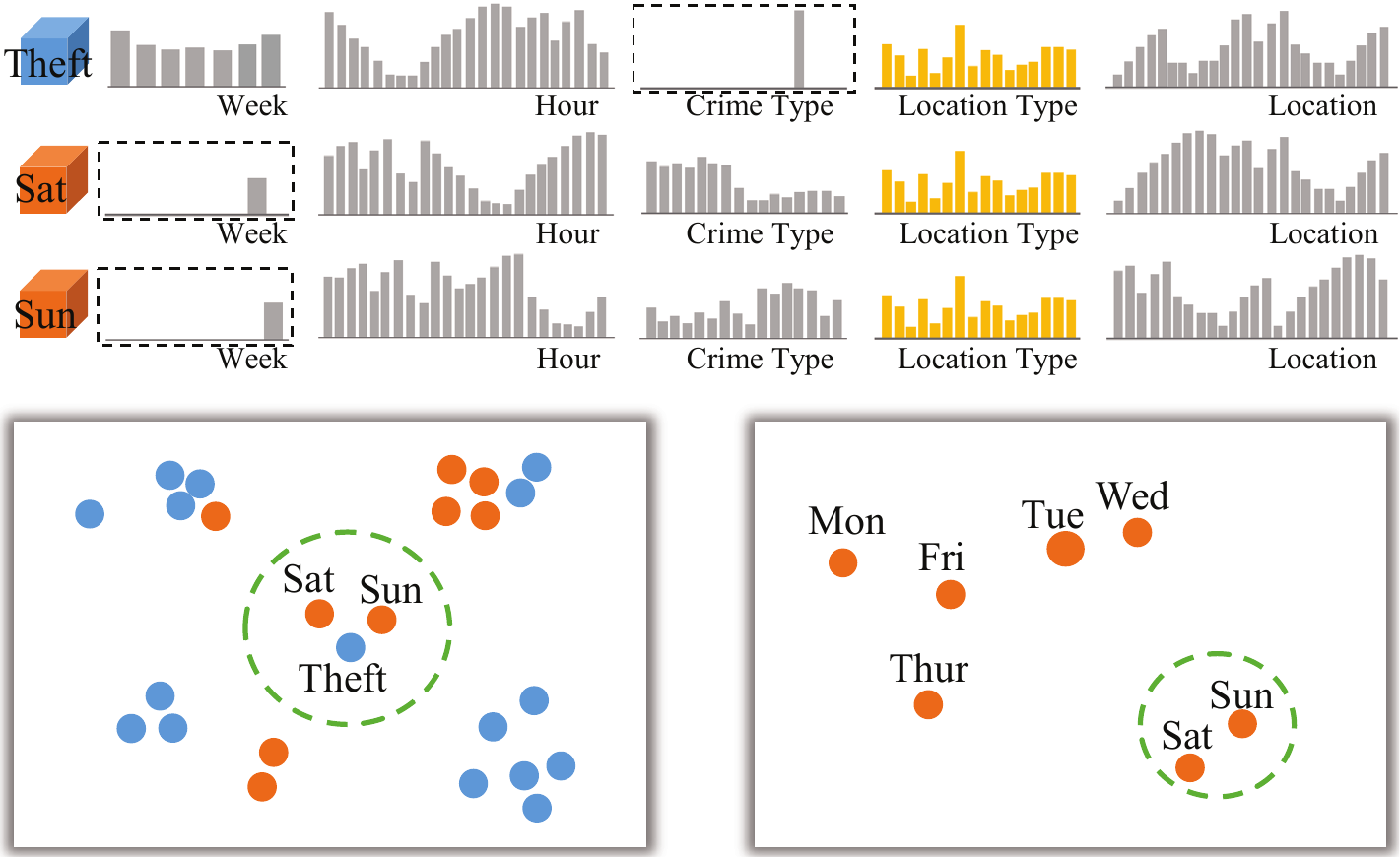}
 \begin{picture}(0,0)
    \put(-65,0){$(a)$}
    \put(60,0){$(b)$}
 \end{picture} 
 \vspace{-0.05 in}
 \caption{Two important design requirements of the SEN that should (a) accommodate arbitrary subsets with  different slicing attributes and features and (b) capture the similarity of subsets on single features. }
 \label{fig:pic2}
\end{figure}
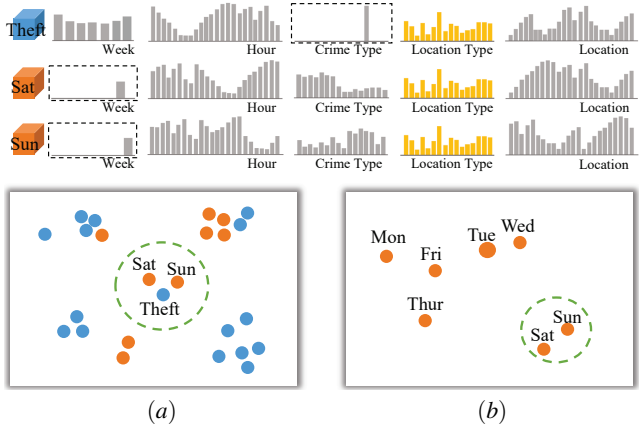

According to the above definition, we can represent a subset as a group of feature vectors. The function of the SEN thus is to take feature vectors of a subset as the input and output its embedding. The nature of subset, however, makes existing embedding techniques \cite{maaten2008visualizing,xie2011m} inapplicable, as follows:

First, existing techniques target on objects with features of the same shape (e.g. number and size). Patterns, however, may exist within arbitrary subsets that have different slicing attributes and features. Figure \ref{fig:pic2}a show such a case, in which we project subsets sliced on Crime-type and Week separately together and exclude features of slicing attributes (marked with dashed borders). Taking them as features will incorrectly increase the similarity of subsets, since their values are zero at most positions. Projections containing diverse subsets involve more interesting patterns. In Figure \ref{fig:pic2}a, the theft subset is adjacent to Saturday and Sunday subsets. We thus speculate theft crimes mainly occur at that time. For applicability, the SEN should be able to deal with subsets with arbitrary slicing attributes (R1). 

Second, existing techniques are designed to capture the overall similarity of target objects across all features. In contrast, subset patterns are often related to few features. For example, in Figure \ref{fig:pic2}b, two week subsets (Saturday and Sunday) have similar values on a feature (see the feature marked in yellow). Existing techniques cannot output similar embeddings for them due to their significant differences on the other features, resulting in missing patterns or finding incorrect patterns when exploring the latent space. To capture the similarity of subsets on single features is necessary for the SEN (R2).

Finally, existing techniques are always restricted by the slow training speed. Moreover, as a self-supervised technique, a well-trained embedding network can only output embeddings for the objects in the training set, different from the offline training of classification models. Therefore, we have to train a SEN on-the-fly after slicing a multi-dimensional data into subsets. A high training efficiency is desired for the network to facilitate the application in a visual analytics environment (R3).

\subsection{Mutli-view Learning-inspired Subset Embedding}
\label{sec:section2.3} 
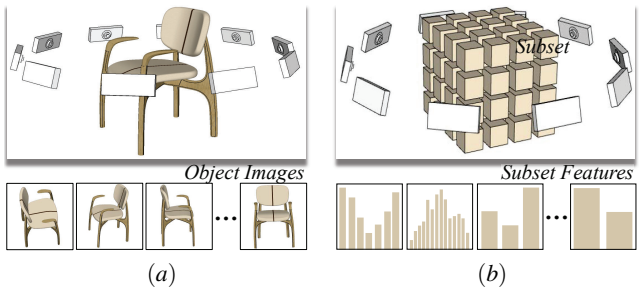
\begin{figure}[htb]
 \centering
 \includegraphics[width=\columnwidth]{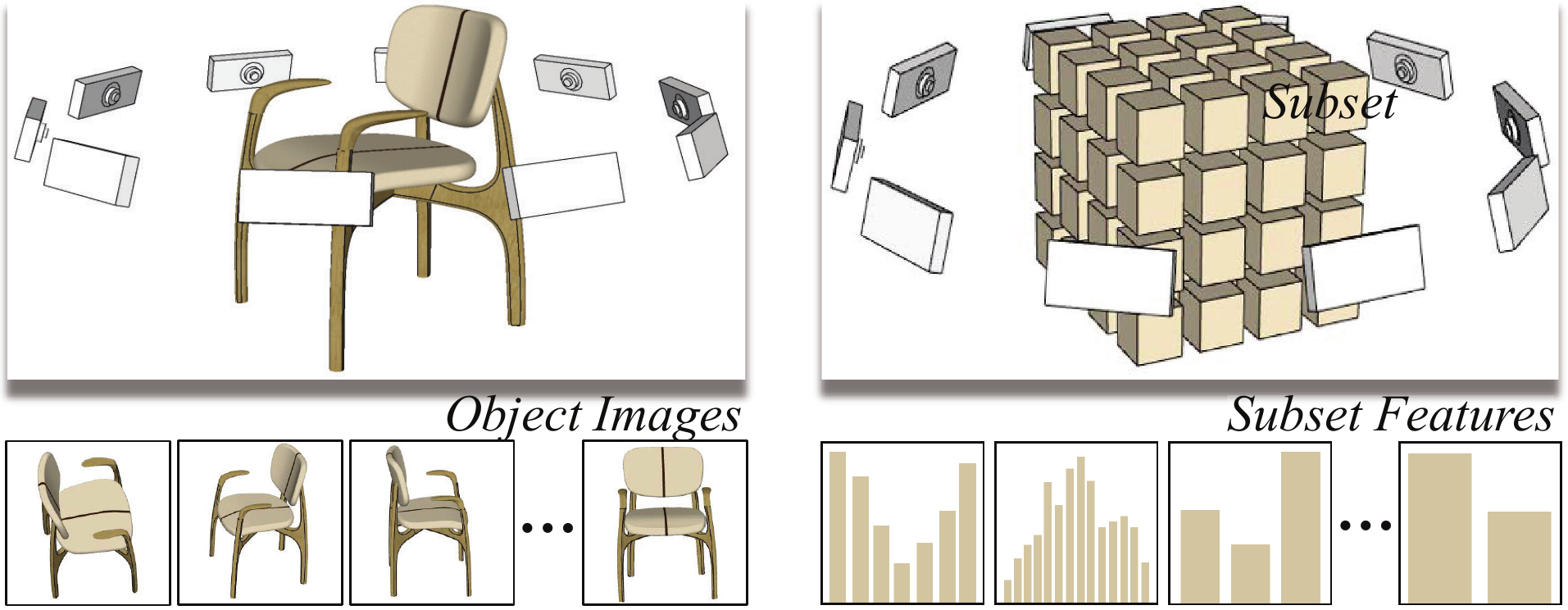}
  \begin{picture}(0,0)
    \put(-65,0){$(a)$}
    \put(60,0){$(b)$}
  \end{picture} 
 \vspace{-0.05 in}
 \caption{Multi-view learning-inspired subset embedding. (a) A multi-view learning model takes multiple views (snapshot) of a target object as the input. (b) We consider a subset as a multi-view object with each feature being a snapshot taken by a virtual camera around the subset.}
 \label{fig:pic3}
\end{figure}

We use the idea of multi-view learning to design the SEN. Multi-view learning is a common kind of machine learning techniques for multi-view objects \cite{li2018survey, Zhao2017Multi, Tang2017survey}. A typical multi-view learning model takes multiple snapshots (views) of an object as the input, as in Figure \ref{fig:pic3}a. Its purpose is taking advantage of the complementary information between views to generate embeddings, thus improving the accuracy of the following object identification or classification tasks. Inspired by multi-view learning, we consider each subset a multi-view object. Each feature can be viewed as a snapshot taken by a virtual camera from different perspectives around the subset, as in Figure \ref{fig:pic3}b. 

\begin{figure*}[htb]
 \centering
 \includegraphics[width=\linewidth]{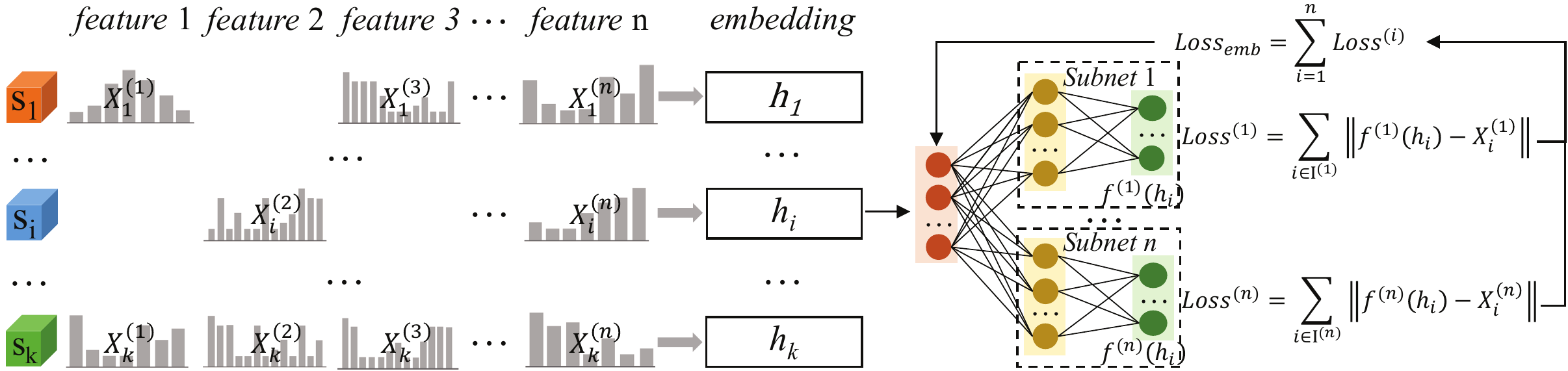}
 
  \begin{picture}(0,0)
    \put(-140,0){$(a)$}
    \put(150,0){$(b)$}
  \end{picture} 
  \vspace{-0.05 in}
 \caption{Subset embedding network. (a) Subsets to be embedded, which are allowed to have different numbers of features. (b) The multi-subnet structure of the SEN. All the subnets share embeddings of subsets as their inputs. }
 \label{fig:pic4}
\end{figure*}
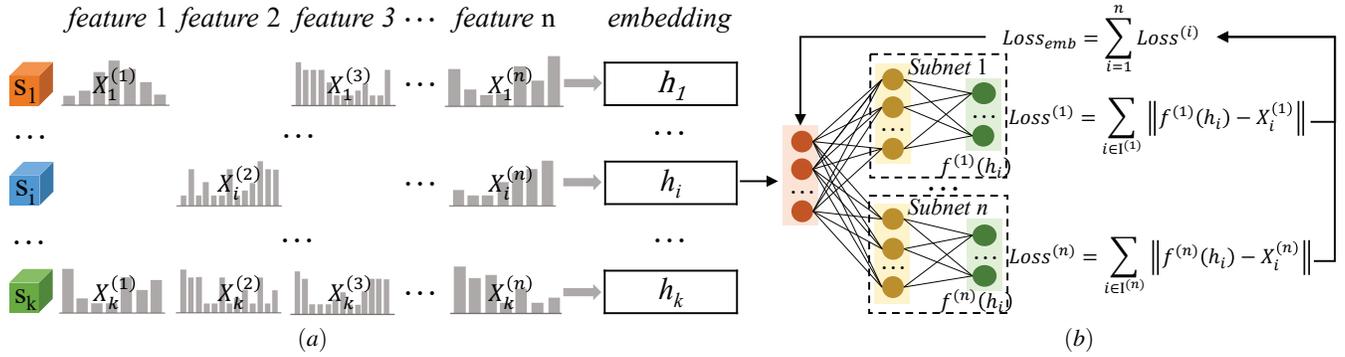

A common characteristic of multi-view learning techniques is they separately treat individual views. Along this line, we propose a SEN of multi-subnet structure, as in Figure \ref{fig:pic4}b. The core idea of the structure is using independent subnets to handle different features and fuse information of different features into final embeddings. In the next section, we will show how the multi-subnet structure satisfies the above three requirements.

\section{Subset Embedding Network}
\label{sec:section3} 

We propose a SEN of multi-subnet structure. Without loss of generality, we allow target subsets to have different numbers of features, as in Figure \ref{fig:pic4}a. The network structure is shown in Figure \ref{fig:pic4}b. The SEN consists of multiple subnets ($f^{(1)}$,$f^{(2)}$,…,$f^{(n)}$), each corresponding to a feature. Let $h_{i}$ be the embedding of subset $S_{i}$. A subnet $f^{(v)}$ takes $h_{i}$ as the input and predicts the $v$th feature vector of $S_{i}$. All the subnets share embeddings of  subsets as inputs. The embeddings are randomly initialized and iteratively updated during the training of the subnets. By using the multi-subnet structure, the network size is linearly proportional to the number of features (the number of subnets). Below we define the losses for updating embeddings and parameters of each subnet.

Let $Loss^{(v)}$ be the loss of the $v$th subnet $f^{(v)}$, we use the sum of losses of all the subnets to update the embeddings, i.e.:

\begin{equation}\label{con:equation2}Loss_{emb} = \sum_{v=1}^{n}Loss^{(v)}  
\end{equation}

\noindent We use the difference between the predicted feature vector $f^{(v)}(h_{i} )$ and the original feature vector  $X_{i}^{(v)}$ as the loss of subnet $f^{(v)}$, i.e.:

\begin{equation}
\label{con:equation1}
Loss^{(v)}=\sum_{i\in I^{(v)}}^{}\left \| f^{(v)}(h_{i} )-X_{i}^{(v)}   \right \| 
\end{equation}

\noindent$I^{(v)}$ represents the set of subsets containing the $v$th feature. Using Figure \ref{fig:pic4}a as an example, $S_{1}$ belongs to $I^{(1)}$, $I^{(3)}$ and $I^{(n)}$, while $S_{i}$ belongs to $I^{(2)}$ and $I^{(n)}$. By introducing $I^{(v)}$, each subset will activate different subnets. That is we only use the losses of subnets corresponding to features owned by a subset to update its embedding. The structure thus is applicable for arbitrary subsets with different numbers of features (R1). 


The multi-subnet structure enables to capture the similarity of subsets on single features. First, embeddings of subsets with few similar features can be similar. Let ($h_{i}$, $h_{j}$) and ($X_{i}^{(v)}$, $X_{j}^{(v)}$) be the embeddings and the $v$th feature vectors of two subsets $S_i$ and $S_j$. When $X_{i}^{(v)}$ and $X_{j}^{(v)}$ are similar, $h_{i}$ and $h_{j}$ should be similar to some extent, as they will go through the same subnet $f^{(v)}$ to obtain similar outputs, i.e. $h_{i}$->$f^{(v)}$-> $X_{i}^{(v)}$, $h_{j}$->$f^{(v)}$-> $X_{j}^{(v)}$. Moreover, subsets with more similar features will have more similar embeddings, as we use the sum of losses of all subnets (Equation \ref{con:equation2}) for updating embeddings (R2).

We implement each subnet as a fully-connected neural network containing a single hidden layer. We fix the length of embeddings to 30, which achieves relatively good performance in most cases. We can conveniently increase the size to reduce the information loss of embedding. Training optimizations, such as learning rate decay, early stop, etc., can be used optionally. The simple structure and few parameters achieve high training efficiency. The network can handle a large number of subsets with multiple features in real-time (Section \ref{sec:section5.3}), enabling easy integration with visualization systems (R3).

Algorithm \ref{alg:SEN} shows the training process of the SEN. The network takes a group of subsets represented as feature vectors as inputs and outputs uniformly-formatted embeddings for them. We randomly initialize embeddings and parameters of subnets (line 1), train subnets with separate losses using Equation \ref{con:equation1} (lines 3-5) and update embeddings using Equation \ref{con:equation2} (lines 6). We will terminate the training when $Loss_{emb}$ (Equation \ref{con:equation2}) no longer significantly decreases over multiple consecutive epochs.

\begin{algorithm}
 \label{alg:SEN}
\caption{Subset Embedding Network Training}
\LinesNumbered 
\KwIn{subsets $\left \{ S_{1},...,S_{k}  \right \}$, $S_{i}=\left \{ X_{i}^{(1)},...,X_{i}^{(n)}  \right \}$ }
Randomly initialize embeddings $\left \{ h_{1},...,h_{k}  \right \}$ and parameters of subnets $\left \{ \theta ^{(1)},...,\theta ^{(n)}   \right \}$ 

\While{not converged}{
    \For{v = 1 : n}{
        Update $\theta ^{(v)}$ with $Loss ^{(v)} = \sum_{i\in I^{(v)}}^{}\left \| f^{(v)} (h_{i} ) - X_{i}^{(v)}  \right \|$ 
    }
    Update $\left \{ h_{1},...,h_{k}  \right \}$ with $Loss_{emb} =  \sum_{v=1}^{n} Loss^{(v)}$ 
}
\KwOut{embeddings $\left \{ h_{1},...,h_{k}  \right \}$ }
\end{algorithm}

\section{Visualization System}
\label{sec:section4} 
We develop a visualization system based on the SEN for exploring multi-dimensional data, detailed below.

\subsection{System Overview}
\begin{figure}[htb]
 \centering
 \includegraphics[width=\linewidth]{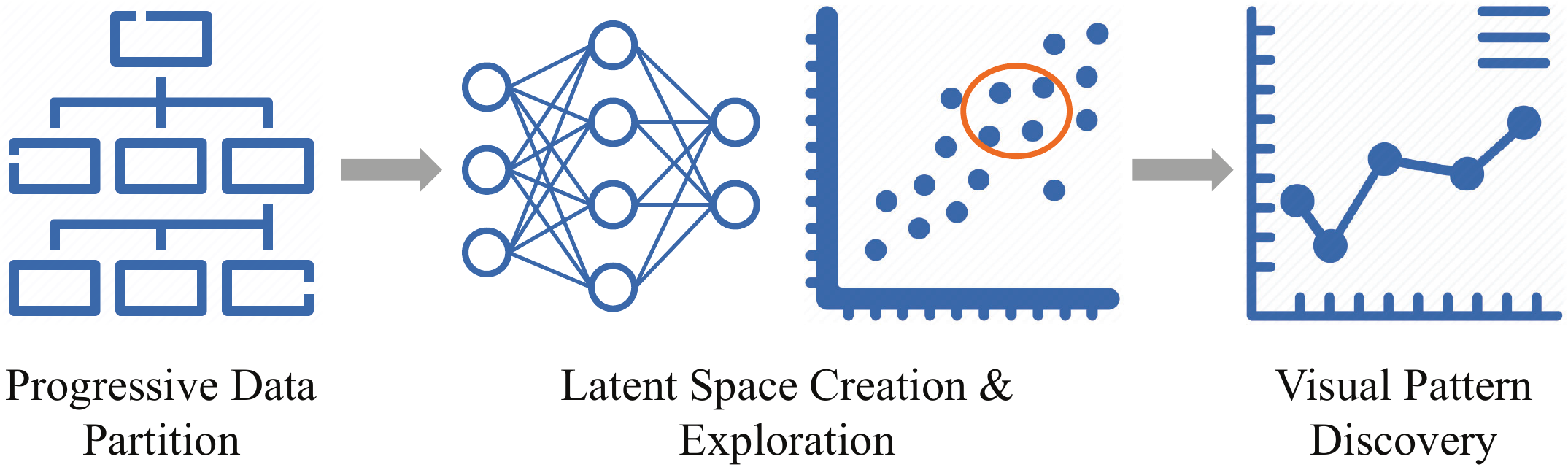}
 \vspace{-0.1 in}
 \caption{The workflow utilized by the visualization system.}
 \label{fig:pic5}
\end{figure}

Following the line of “overview->details->patterns” \cite{shneiderman1996eyes}, we propose an explorative workflow, as in Figure \ref{fig:pic5}. Analysts first partition the given multi-dimensional dataset into subsets (\textbf{Progressive Data Partition}). They then project the subsets according to their embeddings obtained from a SEN trained on-the-fly, and select subsets with specific distributions (e.g. outliers or clusters) in the projection (\textbf{Latent Space Creation \& Exploration}). Analysts finally visualize feature vectors of the selected subsets and observe whether they have consistent values on any features. The more consistent feature values a group of selected subsets have, the more significant patterns they involve (\textbf{Visual Pattern Discovery}).

Figure \ref{fig:pic6} shows the interface of the visualization system that integrates three components, i.e. Exploration Manager (EM), Subset Projector (SP), and Pattern Explainer (PE), to achieve the above workflow, as follows:

\begin{figure*}[htb]
 \centering
 \includegraphics[width=\textwidth]{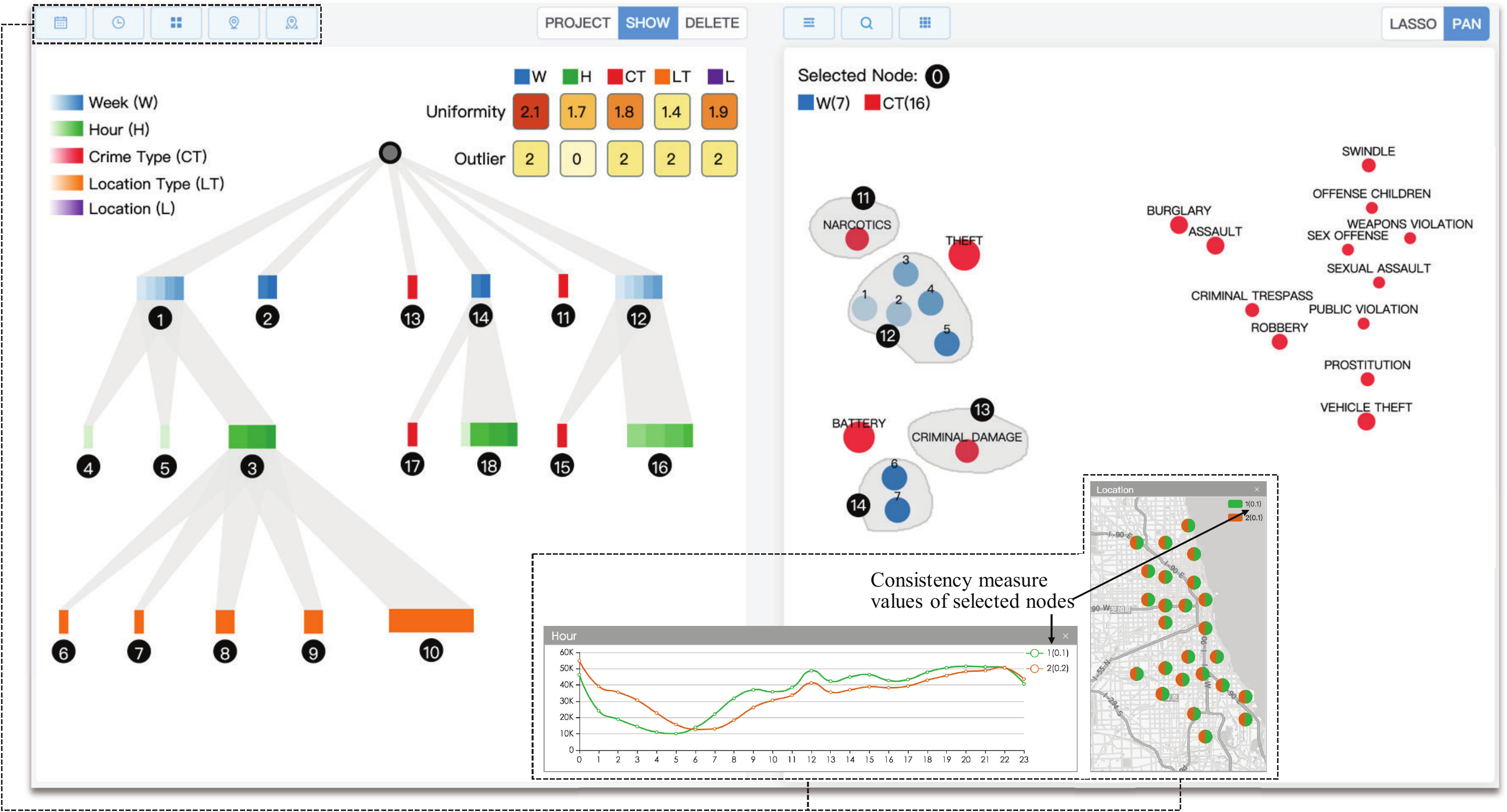}

  \begin{picture}(0,0)
    \put(-150,250){$(a)$}
    \put(92,250){$(b)$}
    \put(-60,80){$(c)$}
  \end{picture}
\vspace{-0.15 in} 
 \caption{The interface of the visualization system that integrates three components, i.e. (a) Exploration Manager (EM), (b) Subset Projector (SP), and (c) Pattern Explainer (PE).}
 \label{fig:pic6}
\end{figure*}
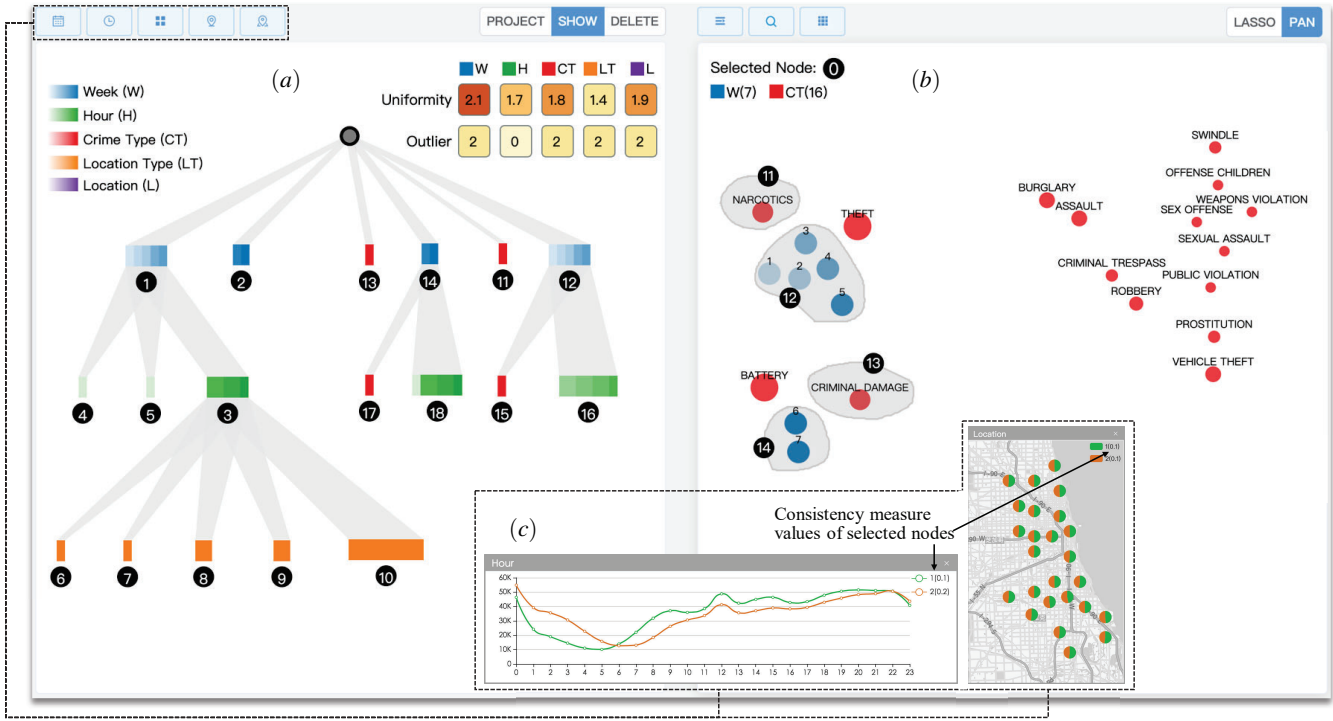

EM implements the progressive data partition that enables analysts to generate a variety of subsets with few operations (step 1), as in Figure \ref{fig:pic6}a. Its main body is a tree. Each node contains subsets selected together from those sliced on its parent. 

SP is responsible for creating the projected latent space, in which analysts select subsets that may involve patterns (step 2), as in Figure \ref{fig:pic6}b. It integrates rich interactions and automatic methods to assist in selecting subsets. 

PE consists of a group of feature cards implemented as classic visualization techniques to show and compare feature values of selected subsets (step 3), as in Figure \ref{fig:pic6}c. It uses a consistency measure (Section \ref{sec:section4.4}) to quantify the significance of the patterns.

\subsection{Progressive Data Partition}
To enable analysts' exploration, we should generate a large number of subsets at the beginning of the exploration. For flexibility, the visualization system should support to generate subsets with any slicing attributes covering any attribute ranges. For efficiency, it is impossible to set slicing attributes and attribute ranges for each subset individually, which will introduce a large number of repetitive operations. We thus propose the progressive data partition to balance the two aspects:

First, we limit to slicing data on an attribute. Therefore, each generated subset has a single slicing attribute. Moreover, we consider subsets selected at a time as a new dataset and allow for partitioning it into subsets again, which actually add a slicing attribute. 

Second, we discretize the value range of the slicing attribute. Specifically, geographical locations are grouped into nominal units, such as states, cities or zones; temporal attributes are discretized on their natural intervals, such as hours, days, or weeks; numerical attributes are divided into equal intervals. We make each generated subset cover a minimum value interval.

\begin{figure}[htb]
 \centering
 \includegraphics[width=\linewidth]{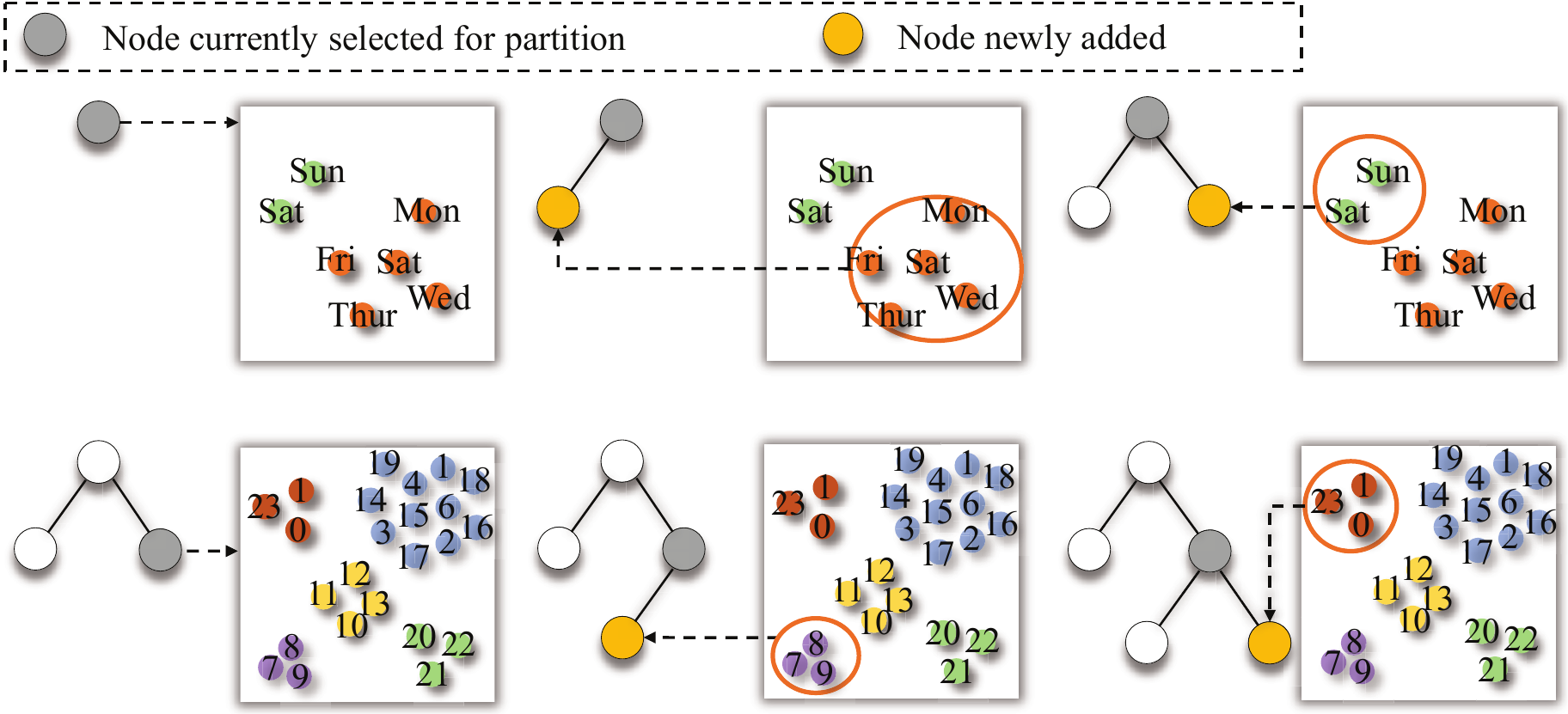}
 
   \begin{picture}(0,0)
    \put(-87,56){$(a)$}
    \put(-8,56){$(b)$}
    \put(74,56){$(c)$}
    \put(-87,3){$(d)$}
    \put(-8,3){$(e)$}
    \put(74,3){$(f)$}
  \end{picture}
 \vspace{-0.05 in} 
 \caption{An example of the progressive data partition. Subsets in a cluster are marked with the same color.}
 \label{fig:pic7}
\end{figure}
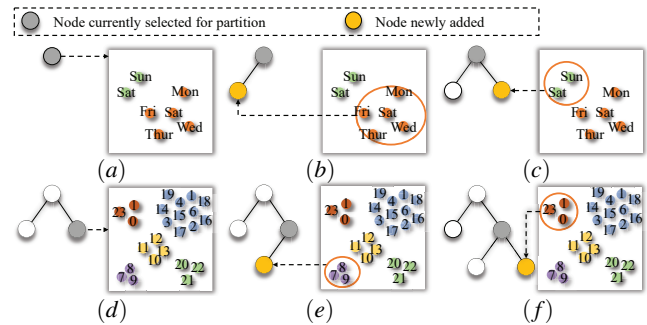

The visualization system utilizes EM to implement the progressive data partition. The main body of EM is a tree, as in Figure \ref{fig:pic6}a. We utilize Figure \ref{fig:pic7} as an example to illustrate how analysts use EM to generate subsets with different slicing attributes. At the beginning of the exploration, there is only the root in the tree, which represents the original dataset. Analysts can select the root and specify “Week” as the slicing attribute to partition it into 7 subsets and project all the generated subsets, as in Figure \ref{fig:pic7}a. They then  select two groups of subsets corresponding to the weekday and the weekend respectively, which form separate child nodes of the root, as in Figure \ref{fig:pic7}b-c. Analysts further select the newly added node that contains subsets of weekend and partition it into 24 subsets on Hour, as in Figure \ref{fig:pic7}d. Each subset thus represents an hour on weekends. They also select two groups of subsets to form two child nodes of the weekend node, as in Figure \ref{fig:pic7}e-f. 

Analysts can generate arbitrary subsets they want by repeating the above process. Moreover, the progressive process balances the flexibility and efficiency of the subset generation. First, it simplifies interactive operations. Analysts can simply choose a slicing attribute to generate subsets without the needs of setting the attribute range for each subset individually. Second, it makes the implementation of the visualization system easy. The system only needs to support a type of operations, i.e. slicing a dataset into subsets on a user-specified attribute. Finally, each projected subset covers a unit of the value range of the slicing attribute, which avoids generating a large number of logically-unrelated subsets, making identified patterns more explainable.

In EM, each tree node consists of multiple colored rectangles, as in Figure \ref{fig:pic6}a. Each rectangle represents a subset that covers a single unit of the value range of the slicing attribute. The more subsets a node contains, the longer the node will be. We assign a globally unique color to an attribute, see the legend in Figure \ref{fig:pic6}a. The color of a rectangle indicates the slicing attribute selected at the current round. We can understand all the slicing attributes of subsets of a node by tracing colors of nodes along the path from the root to the node. For example, in Figure \ref{fig:pic6}a, \#1 has a single slicing attribute, i.e. Week, while \#3 has two slicing attributes, i.e. Week and Hour. For attributes with continuous attribute ranges (e.g. Week and Hour), the transparency of rectangles (subsets) gradually decreases as the attribute value increases. Rectangles of categorical attributes have the same transparency. 

EM will display two measure values, i.e. uniformity and the number of outliers \cite{seo2004rank} for each marginal distribution of the selected node (see the colorful rectangles in the upper right corner of Figure \ref{fig:pic6}a, which is showing ten measure values of the root node on five attributes). The two measures provide important guidance for selecting slicing attributes, especially helpful when the dataset contains many attributes. A larger or smaller value may relate to potential patterns. Therefore, analysts can slice the node on the corresponding attribute.

\subsection{Latent Space Creation \& Exploration}
We extract features for sliced subsets, project them according to their embeddings obtained from a SEN trained on-the-fly, and select parts of subsets that may involve patterns, as follows:

\textbf{Feature Extraction}. We calculate the distribution of the number of records on an attribute as a feature for a subset, as in Figure \ref{fig:pic2}. It is also possible to add/remove features, or use other features for each subset. We exclude features of slicing attributes (the reason has been explained in Section \ref{sec:section2.2}). Using the five-dimensional Chicago crime dataset as an example (Figure \ref{fig:pic1}), each 1-dimensional subset has four features, each 2-dimensional subset has three features, and so on. 

\textbf{Representation Learning \& Projection.} We train a SEN to obtain uniformly-formatted embeddings of the subsets. We then project the subsets onto a 2-dimensional plane according to their embeddings. We choose t-SNE \cite{maaten2008visualizing} to generate the projection. Other dimensionality reduction techniques can also be used.

\textbf{Subset Selection.} Each point in the generated projection represents a subset, as in Figure \ref{fig:pic6}b. The size of a point encodes the number of records contained in the subset. Analysts select subsets in the projection, which are mapped together as a child of the selected node in the tree. Common interactions, such as zoom, pan, lasso, etc., are supported by SP. 

There are three ways to select subsets in the SP, as follows:

\textbf{1)} Analysts can freely select subsets according to the distribution of the projected subsets. Significant clusters or outliers are possible candidates for selection. 

\textbf{2)} Analysts can highlight projected subsets within specified attribute ranges and select the highlighted subsets exclusively. For example, in Figure \ref{fig:pic9}c, we highlight seven subsets (sliced on Hour). A common explorative strategy is to change the queried attribute range and observe the distribution of the highlighted subsets in the projection before selecting subsets. Having found any interesting distributions (e.g. clusters or outliers), analysts can select them for further in-depth analysis. 

\textbf{3)} Analysts can use the clustering function to divide projected subsets into groups according to their embeddings automatically. Many clustering algorithms, such as K-means, hierarchical clustering, density-based clustering (no need to specify the number of clusters), etc., are integrated. Analysts can set parameters, such as the number of clusters, according to their prior assumptions. For example, we can divide seven subsets sliced on Week into two clusters, considering weekday and weekend are naturally different on many features. They can also interactively adjust the parameter during the exploration. Clustering results, whether expected or not, provide cues for the subset selection. A group of links will appear on the right of the projection after clustering. Each link corresponds to a cluster of subsets, as in Figure \ref{fig:pic9}d. We can click a link to highlight the corresponding subsets.

\subsection{Visual Pattern Discovery}
\label{sec:section4.4} 
We use PE to visualize the feature vectors of subsets of different nodes in the tree, as in Figure \ref{fig:pic6}c. Feature vectors of subsets in a node are aggregated and visualized as a whole in each feature card. Each feature card corresponds to a feature and is implemented as a classical visualization technique suitable for the feature. For example, line chart is for showing temporal patterns; bar chart is for showing categorical patterns; thematic map (pie chart + map) is for showing spatial patterns. Analysts can select multiple nodes and add them into a feature card for comparison. Figure \ref{fig:pic6}c contain two nodes respectively, in which each visual item (line/bar/sector) reflects the aggregated feature vector of all the subsets in a node. A feature card will appear when analysts click on the corresponding button in the upper left corner of the system, marked with dashed borders in Figure \ref{fig:pic6}. 

\begin{figure}[htb]
 \centering
 \includegraphics[width=\linewidth]{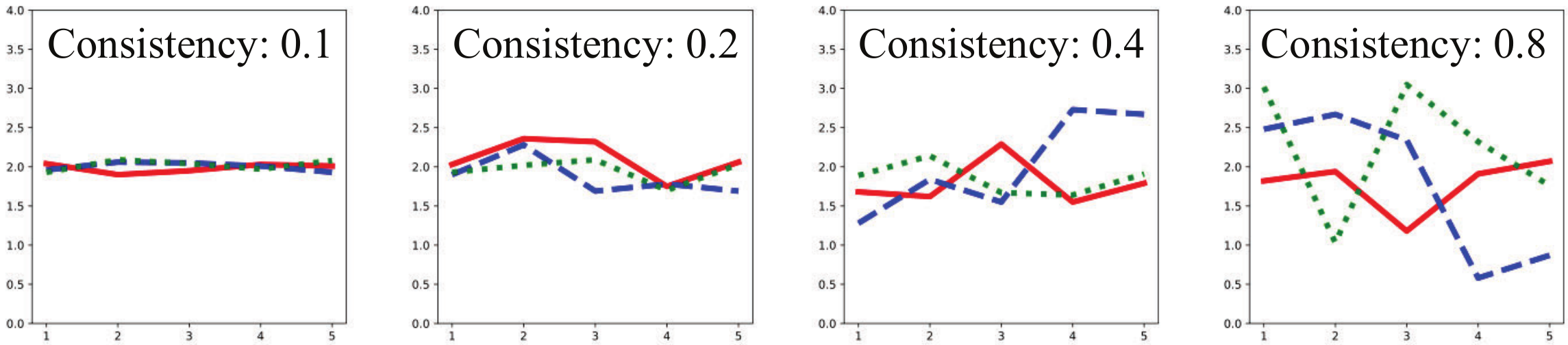}
 \vspace{-0.1 in}
 \caption{Four groups randomly-generated vectors with different consistency measure values.}
 \label{fig:pic8}
\end{figure}

We propose a measure to describe the consistency of feature vectors of subsets contained in a node using $\frac{1}{D} \sum_{d=1}^{D}\sigma (d)$, where D represents the size of the feature vector, and $\sigma (d)$ represents the standard deviation of feature vector values at the $d$th positions of all subsets. A lower measure value indicates more consistent feature vectors (as in Figure \ref{fig:pic8}), i.e. more significant subset patterns. We visually encode the measure value for each node in feature cards (see numbers in legends in Figure \ref{fig:pic6}c). 

\subsection{Implementation}
The visualization system has a client-server architecture. The SEN is at the backend, which receives requests from the frontend (written in JavaScript using D3 library \cite{bostock2011d3}) and sends back embeddings. The Flask is used to transfer parameters between the front and back ends. We also build an OLAP index, i.e. data cube, to speed up the feature extraction from the original data.
\section{Evaluation}
\label{sec:section5} 
We conduct case studies and quantitative experiments to demonstrate the applicability of our approach in actual scenarios.

\subsection{Case Studies}
\begin{figure*}[htb]
 \centering
 \includegraphics[width=\textwidth]{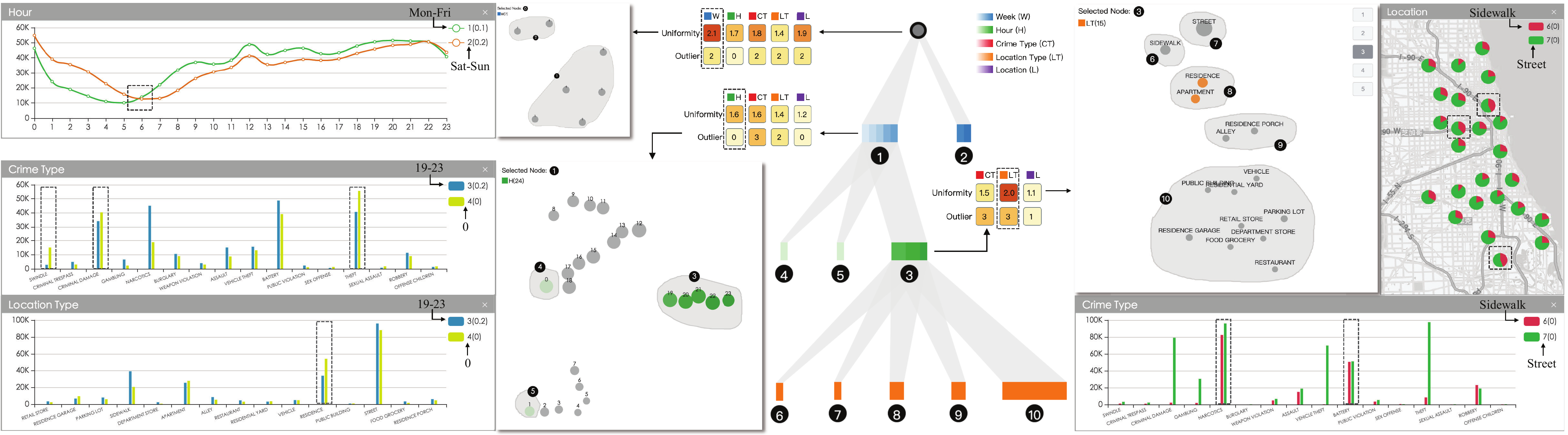}

  \begin{picture}(0,0)
    \put(50, 140){$(a)$}
    \put(-70, 140){$(b)$}
    \put(-165,140){$(b1)$}
    \put(-25, 90){$(c)$}
    \put(-165,90){$(c1)$}
    \put(-165,50){$(c2)$}
    \put(170,140){$(d)$}
    \put(207,50){$(d1)$}
    \put(207,140){$(d2)$}
  \end{picture}
\vspace{-0.2 in} 
 \caption{Explorative process of the first category of tasks. The exploration begins at the root node (a), then we gradually add slicing attributes to generate projections (b-d) and identify patterns (b1-d2).}
 \label{fig:pic9}
\end{figure*}
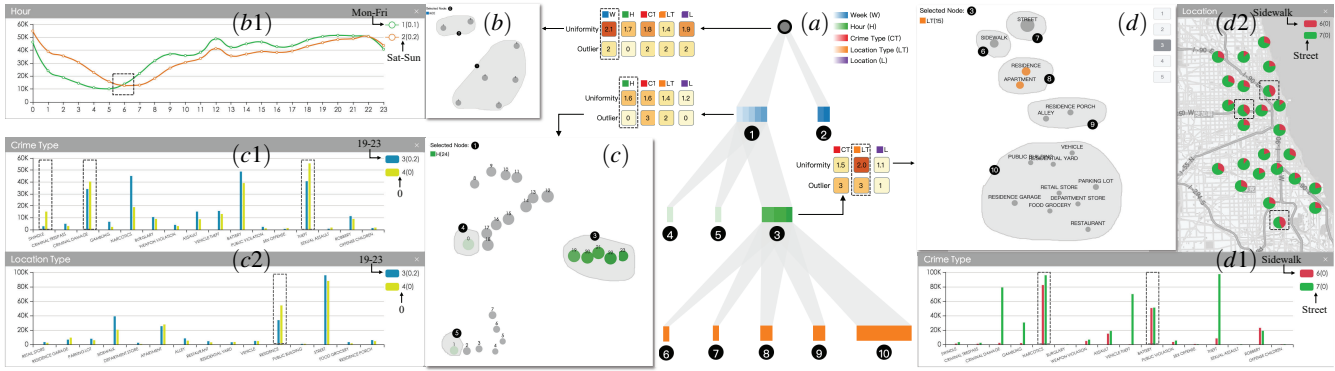

We develop a visualization system on the Chicago crime dataset \cite{chicagocrime} (about 6M records) and perform two categories of tasks to identify patterns at group and individual levels respectively:

\textbf{Group Segmentation.} We partition data into subsets on gradually-increasing slicing attributes. For each partition, we select groups of subsets with similar feature values. Each group thus indicates a kind of crime patterns. The exploration process is shown in Figure \ref{fig:pic9}a and detailed below.

We choose Week with the highest uniformity value as the slicing attribute to partition the data into 7 subsets. We project the subsets (Figure \ref{fig:pic9}b) using t-SNE (perplexity is set to 20, while other hyperparameters are kept as default settings) and find two obvious clusters, i.e. weekday (\#Mon- \#Fri) and weekend (\#Sat-\#Sun). We select them as separate tree nodes (\#1 and \#2) and determine patterns using feature cards. We find crime patterns in a feature card (Figure \ref{fig:pic9}b1), i.e. more crimes occur in the daytime on weekdays (green line), while more occur at night on weekends (orange line). 

We further partition \#1 into 24 hour subsets to understand finer crime patterns of weekdays. We focus on subsets of night (19:00-1:00), when crimes occur intensively. We highlight the corresponding subsets, which form 3 clusters in the projection, as in Figure \ref{fig:pic9}c. We select them as three separate tree nodes (\#3-\#5) and find several interesting patterns. For example, some kinds of crimes (marked with dashed borders in Figure \ref{fig:pic9}c1) occur more often at midnight (0:00) than in the evening (19:00-23:00), while residences are the main locations (marked with dashed borders in Figure \ref{fig:pic9}c2).

We finally partition \#3 on Location-type. Each subset thus represents a location-type of weekday evening. The subsets are automatically divided into 5 clusters, as in Figure \ref{fig:pic9}d. An interesting finding is two roadway-related subsets, i.e. sidewalk and street, are in two clusters. By selecting them and visualizing their features, we find they are drastically different on features of crime-type and location. Two criminal types (marked with dashed borders in Figure \ref{fig:pic9}d1) and three locations (marked with dashed borders in Figure \ref{fig:pic9}d2) are related to sidewalk, while crimes occurring on street are more diverse and evenly distributed throughout the city.

\begin{figure}[htb]
 \centering
 \includegraphics[width=\linewidth]{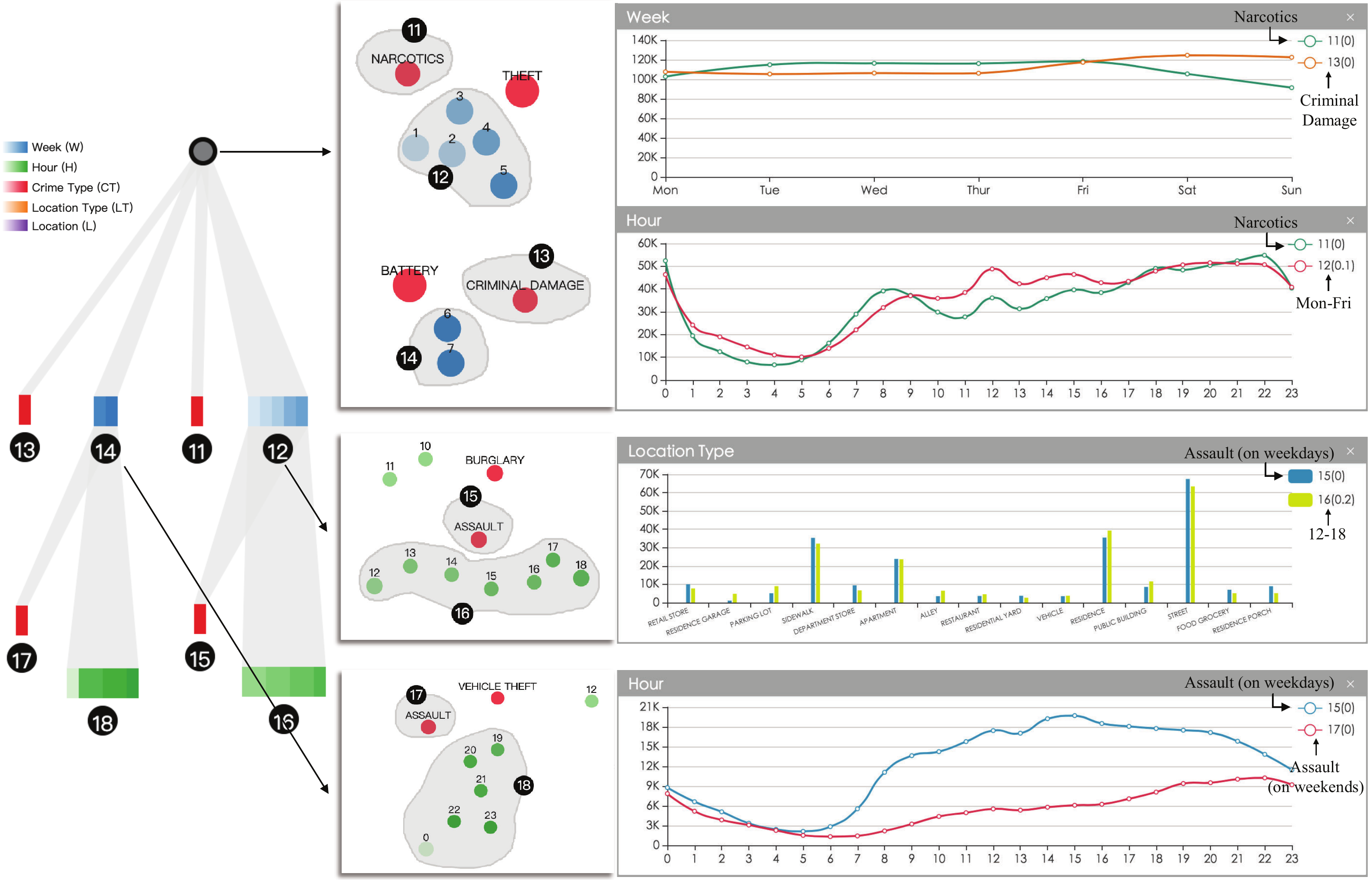}
 
  \begin{picture}(0,0)
    \put(-25,158){$(a)$}
    \put(55,158){$(a1)$}
    \put(55,122){$(a2)$}
    \put(-25,82){$(b)$}
    \put(55,82){$(b1)$}
    \put(-25,41){$(c)$}
    \put(55,41){$(c1)$}
  \end{picture}
  \vspace{-0.1 in}
 \caption{Explorative process of the second category of tasks. (a) Projecting 16 crime-type subsets and 7 week subsets together. (b-c) Projecting 16 crime-type subsets and 24 hour subsets sliced on weekday and weekend selected in (a) respectively. (a1-c1) Feature cards for determining patterns.}
 \label{fig:pic10}
\end{figure}
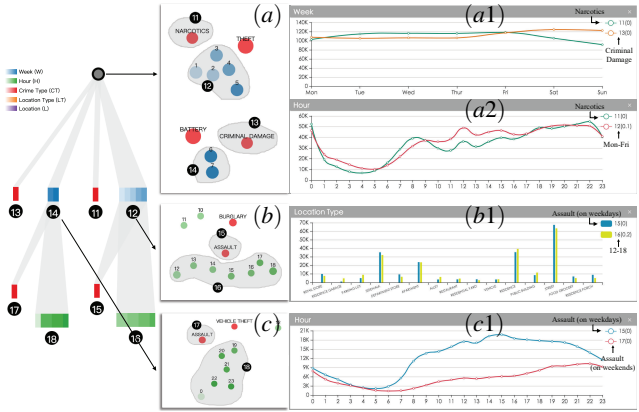

\textbf{Individual Relationship Identification.} We project subsets of crime types and those sliced on different temporal attributes (Week and Hour) together. The purpose is to know when specific kinds of crimes intensively occur, as follows:

We first project the 16 crime-type subsets with those sliced on Week, as in Figure \ref{fig:pic10}a. We find weekday subsets and weekend subsets are close to different crime-type subsets and select them as tree nodes. The finding indicates that narcotics and criminal damage crimes occur more often in different periods of week, as in Figure \ref{fig:pic10}a1. Their similar feature values (as in Figure \ref{fig:pic10}a2) demonstrate the correctness of the projection. We also find similar patterns on theft and battery (omitted for limited space).

We partition \#12 (weekday) and \#14 (weekend) on Hour respectively to form two groups of hour subsets. We project 16 crime-type subsets with each group of hour subsets together, as in Figure \ref{fig:pic10}b-c. We find assault subsets (\#15 and \#17) in the two projections are adjacent to afternoon subsets (12:00–18:00) and evening subsets (19:00–0:00) respectively. Figure \ref{fig:pic10}b1 proves the correctness of the projection, in which the assault subset (\#15) and the selected afternoon subsets (\#16) have similar values on a feature. The observation indicates that assault crimes occur more often in the afternoon on weekdays, while assault crimes occur more often in the evening on weekends, as in Figure \ref{fig:pic10}c1.

\subsection{Quantitative Experiments}
We assess whether SEN can accurately encode single-feature similarity of subsets through two experiments.

\subsubsection{Experiment Design}
 We collect 5 open multi-view datasets for the two experiments, including Handwritten Digits \cite{handwritten}, ORL \cite{orl}, PIE \cite{pie}, Caltech 101-7 \cite{caltech101-7} and BBCSport \cite{bbcsport}. We treat each record of the datasets as a subset, which has multiple features and a label that indicates its category. The records of the same category have similar values on corresponding features.


For each dataset, we replace several features of records with random numbers. This step makes parts of features that are originally similar different, thus simulating the cases that subsets (records) have different numbers of similar features. We train SENs to obtain 30-dimensional emebddings of the records and evaluate the accuracy from the following two aspects: 

\textbf{Pattern Encoding Accuracy}. We use K-means to divide the records of each dataset into clusters according to their embeddings. The number of clusters is set to the actual number of categories of the dataset. We quantify the differences between the actual clusters and predicted clusters using three indicators, i.e. \textbf{(i)} Accuracy (ACC) \textbf{(ii)} Normalized Mutual Information (NMI) \textbf{(iii)} Adjusted Rand Index (ARI) (we match a predicted cluster and a actual one if they have a large number of common records). The value ranges of the three indicators are [0, 1], with 0 and 1 representing the lowest and highest accuracy. A larger value indicates a better match of predicted and actual clusters.

\textbf{Visual Perception Accuracy}. We use t-SNE \cite{maaten2008visualizing} to project records according to their embeddings and evaluate the visual separability of records of different categories in the projection. Two indicators are used: \textbf{(i)} Silhouette Coefficient (SC) and \textbf{(ii)} Calinski-Harabasz Index (CHI). A larger value indicates higher separability, i.e. higher visual perception accuracy.

We compare SEN with t-SNE \cite{maaten2008visualizing} and m-SNE \cite{xie2011m}. We connect all features (vectors) of a record as the input of t-SNE, while m-SNE, as a multi-view dimensionality reduction technique, has the same input format as SEN. For pattern encoding accuracy, all the three techniques output 30-dimensional embedding vectors. For visual perception accuracy, t-SNE and m-SNE outputs 2-dimensional vectors for projection directly, while we project 30-dimensional embedding vectors of SEN through dimensionality reduction (consistent with the actual ``first embedding then projection'' workflow of SEN). We set perplexity to 20 and keep other hyperparameters as default settings for t-SNE and m-SNE.

\subsubsection{Experiment 1}
We first choose four datasets, replace half of features for each dataset, and calculate indicator values for embeddings obtained using the three techniques. As in Table \ref{tab:table1}, we find SEN has the highest values in most cases (marked in red), which indicates general higher pattern encoding accuracy and visual perception accuracy.

\begin{table}[htb]
\center
\caption{Experiment results on four datasets with half of features replaced. The highest indicators are marked in red. We repeat the experiment five times, and numbers in parentheses are variances.}
\label{tab:table1}

\resizebox{\linewidth}{23mm}{
\begin{tabular}{cc|cccc}

\hline
Method & Measure & BBCSport & Caltech-7 & ORL & PIE\\
\hline
\multirow{5}{*}{SEN} & ACC & \textcolor{red}{0.719} (0.097) & \textcolor{red}{0.599} (0.059) & \textcolor{red}{0.630} (0.036) & \textcolor{red}{0.470} (0.144)\\
& NMI & \textcolor{red}{0.535} (0.057) & \textcolor{red}{0.429} (0.065) & \textcolor{red}{0.612} (0.043) & \textcolor{red}{0.439} (0.188)\\
& ARI & \textcolor{red}{0.497} (0.087) & \textcolor{red}{0.354} (0.068) & \textcolor{red}{0.407} (0.055) & \textcolor{red}{0.231} (0.161)\\
& SC & \textcolor{red}{0.214} (0.010) & 0.034 (0.055) & \textcolor{red}{0.009} (0.039) & \textcolor{red}{-0.078} (0.090)\\
& CHI & \textcolor{red}{158.788} (31.466) & 38.825 (16.290) & \textcolor{red}{7.792} (3.143) & \textcolor{red}{5.117} (3.266)\\
\hline
\multirow{5}{*}{m-SNE} & ACC & 0.437 (0.079) & 0.537 (0.179) & 0.349 (0.009) & 0.303 (0.006)\\
& NMI & 0.173 (0.086) & 0.395 (0.250) & 0.240 (0.045) & 0.183 (0.039)\\
& ARI & 0.142 (0.087) & 0.320 (0.225) & 0.083 (0.007) & 0.037 (0.010)\\
& SC & -0.014 (0.027) & \textcolor{red}{0.064}  (0.154) & -0.191 (0.010) & -0.185 (0.012)\\
& CHI & 15.154 (17.449) & 48.553 (51.837) & 1.285 (0.683) & 0.987 (0.233)\\
\hline
\multirow{5}{*}{t-SNE} & ACC & 0.562 (0.016) & 0.528 (0.097) & 0.508 (0.092) & 0.346 (0.071)\\
& NMI & 0.374 (0.022) & 0.359 (0.131) & 0.434 (0.158) & 0.342 (0.168)\\
& ARI & 0.308 (0.021) & 0.282 (0.119) & 0.252 (0.117) & 0.122 (0.095)\\
& SC & 0.150 (0.022) & -0.007 (0.104) & -0.074 (0.119) & -0.146 (0.034)\\
& CHI & 112.595 (11.819) & \textcolor{red}{48.904} (27.649) & 6.809 (5.213) & 1.229 (0.392)\\
\hline
\end{tabular}}
\end{table}

\subsubsection{Experiment 2}
\begin{figure}[htb]
 \centering
 \includegraphics[width=\linewidth]{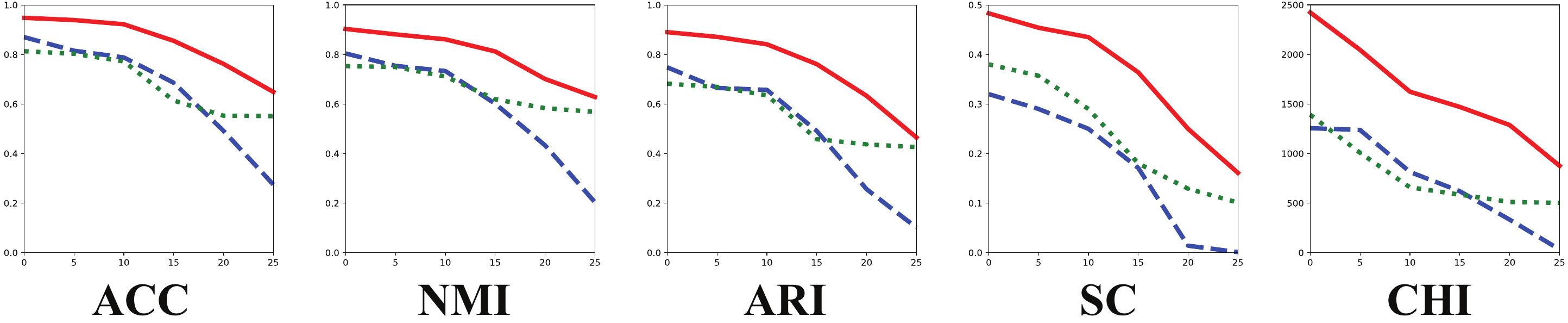}
 \caption{Indicator values of three techniques, i.e. SEN (red lines), m-SNE (blue lines), tSNE (green lines), under different numbers of replaced features. }
 \label{fig:pic11}
\end{figure}

We conduct the experiment using the Handwritten Digits dataset that contains ten categories. We split each record's 649 attributes into 30 features (29 20-dimensional features and 1 69-dimensional feature). Figure \ref{fig:pic11} shows the experiment results. We find indicator values decrease as the numbers of replaced features increase. However, the downward trend of SEN (red lines) is slower than those of m-SNE (blue lines) and t-SNE (green lines). Moreover, SEN has higher values on all the five indicators at any number of replaced features.

\begin{figure}[htb]
 \centering
 \includegraphics[width=\linewidth]{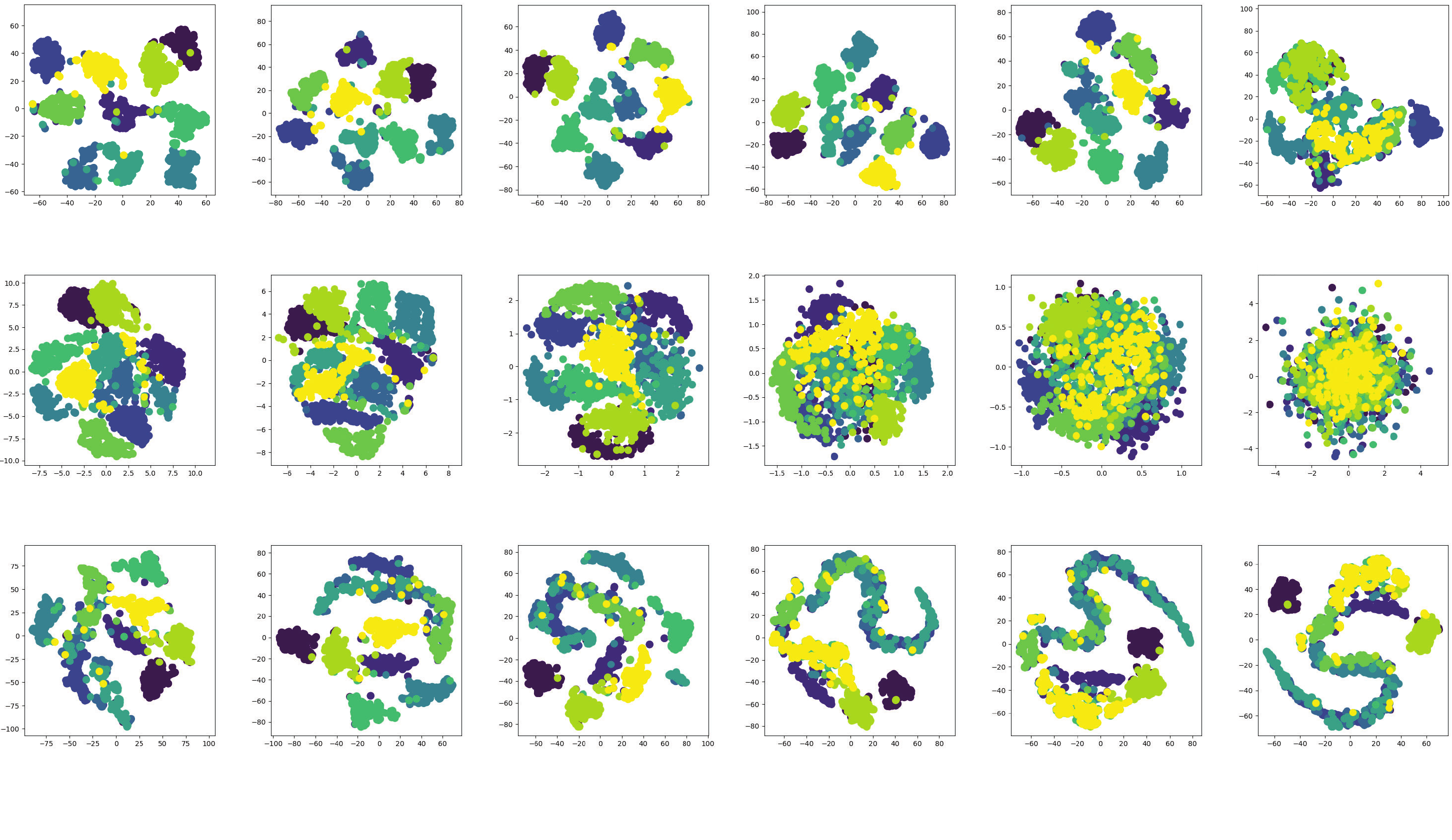}
 
  \begin{picture}(0,0)
    \put(-13,108){$(a)$ SEN}
    \put(-16,63){$(b)$ m-SNE}
    \put(-14,17){$(c)$ t-SNE}
    \put(-107,5){0/30}
    \put(-67,5){5/30}
    \put(-28,5){10/30}
    \put(13,5){15/30}
    \put(53,5){20/30}
    \put(95,5){25/30}
  \end{picture}
  \vspace{-0.1 in}
 \caption{Projections of Handwritten Digits dataset with 0-25 replaced features. From left to right, numbers of replaced features gradually increase.}
 \label{fig:pic12}
\end{figure}
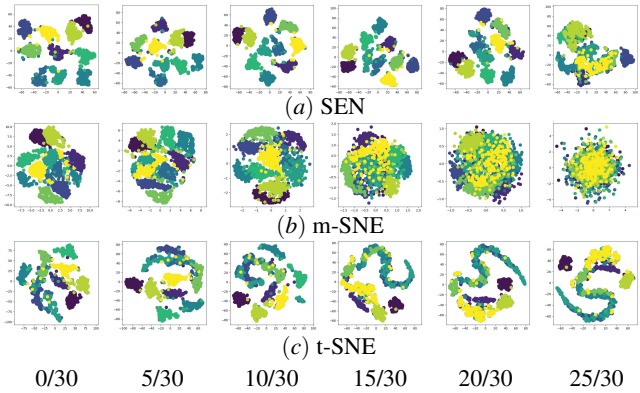

We further visualize projections obtained in the above experiment, as in Figure \ref{fig:pic12}. Each point represents a record (subset) with the color indicating the category. We find when only a small number of features are replaced, most categories can be identified in the projections (see the two leftmost columns). However, as the number of replaced features increases, many categories are mixed in the projections of m-SNE and t-SNE (see the two rightmost columns in the last two rows). In contrast, we can identify more categories in the projections of SEN, even when few features are retained (see the two rightmost projections in the first row).

\subsection{Efficiency Assessment}
\label{sec:section5.3} 
We finally test the training speed of SEN using randomly-generated records (subsets).  We choose two independent variables, i.e. $|subsets|$ and $|features|$. The experiment is conducted on a GPU server (XEON E5-2680, 196G, 2080Ti). 

As in Table \ref{tab:table2}, the SEN can handle hundreds of subsets with multiple features in a short time. Moreover, the longest training time is 7.5s, which is still an acceptable time cost for most visualization systems. Experiment results show high efficiency of the SEN.

\begin{table}[htb]
\center
\caption{Training time (seconds) of the SEN under different numbers of features and subsets. We repeat the experiment five times and numbers in parentheses are variances.}
\label{tab:table2}
\resizebox{\linewidth}{8mm}{
\begin{tabular}{c|c|c|c|c|c}
\hline
\diagbox{$\left | subsets \right | $}{$\left | features \right | $} & 10 & 15 & 20 & 25 & 30\\
\hline
100 & 1.9 (0.18) & 2.9 (0.25) & 4.5 (0.36) & 5.8 (0.43) & 7.2 (0.52)\\
\hline
300 & 1.9 (0.25) & 3.0 (0.27) & 4.7 (0.32) & 6.1 (0.41) & 7.3 (0.54)\\
\hline
500 & 2.0 (0.21) & 3.1 (0.31) & 4.9 (0.41) & 6.2 (0.45) & 7.5 (0.55)\\
\hline
\end{tabular}}
\end{table}
\section{Discussion and Limitations}
\label{sec:section6} 
We discuss two important but easily overlooked aspects, which may affect the application of our approach in actual scenarios.

There is inevitably uncertainty in the projection. To explain this point, consider three subsets A, B, and C. Specifically, A and B have similar values on a feature, A and C have similar values on another one, while B and C are different on all features. In that case, the embedding similarity between B and C is uncertain. They can be similar (as B and C are similar to A on single features) or not (as B and C are completely different). We hope the visualization system can eliminate the uncertainty. The system provides rich interactions and automatic methods to assist in selecting subsets in the projection, and provides feature cards and a consistency measure to visually determine the existence of patterns within selected subsets, thus improving the interpretability and faithfulness of the identified patterns.

As an AI model, the SEN inevitably involves hyperparameters, such as the size of embeddings, the number of neurons in different layers, etc. The black-box nature makes the optimization of these hyperparameters difficult. The positive aspect is we only use ordinary fully-connected neural network with common hyperparameters. The simple structure reduces the tuning difficulty. We only assume developers of the SEN have basic knowledge about neural network. They can always obtain a well-performing model through a small number of trials. The simple network structure and few parameters also achieve fast training.

\section{Related Work}
\label{sec:section7} 
We discuss related works from the following two aspects that are related to the two contributions of the paper.

\subsection{Data Embedding}
DR is the most common embedding technique for multi-dimensional data. PCA \cite{wold1987principal} projects data along the directions with maximum variance. MDS \cite{borg2005modern} preserves pairwise Euclidean distances during the projection. ISOMAP \cite{tenenbaum2000global} changes the Euclidean distance of MDS to geodesic distance, thus enabling to capture nonlinear manifold structures in high-dimensional space. t-SNE \cite{maaten2008visualizing} allows objects that are close in high-dimensional space to be projected together with high probability in low-dimensional projections. Other important DR techniques include LLE \cite{roweis2000nonlinear}, LE \cite{belkin2002laplacian}, LTSA \cite{zhang2004principal}, etc. Many works exist for explaining the DR results \cite{fujiwara2019supporting, liao2017cluster, faust2018dimreader, chatzimparmpas2020t}. All these techniques, however, are to maintain global relationships \cite{spathis2019interactive} and cannot be used to explore subset patterns that are often associated with few features. 

There are many DR techniques with special objective functions \cite{liu2016visualizing, sacha2016visual, espadoto2019towards}. Wang et al. \cite{wang2017perception} propose a DR algorithm that maximizes inter-class distances. Zhang et al. \cite{chen2015uncertainty} offer a technique that can reflect the similarity of target objects on both statistic metrics and distributions. Fujiwara et al. \cite{fujiwara2019incremental} propose a DR technique for streaming data, which can maintain the mental map of analysts for record or dimension changes. Liu et al. \cite{liu2020uncovering} propose a response function preserving algorithm that generates projections showing patterns related to single response functions. SEN is of this category with maintaining the similarity of objects on single features as the objective function.

Applying the neural network in data embedding is a recent trend. Hinton and Salakhutdinov \cite{hinton2006reducing} implement an autoencoder-based embedding technique that achieves better performance than PCA. Mikolov et al. \cite{mikolov2013distributed} propose the famous word2vec that embeds words according to their co-occurrence in a document set. Many similar techniques, such as cite2vec \cite{berger2016cite2vec}, location2vec \cite{zhu2019location2vec}, poi2vec \cite{an2017poi2vec}, etc. have been proposed. These techniques, however embed objects using their co-occurring frequency in the dataset, unable to be used to explore subset patterns.

The SEN is inspired by multi-view learning, as in Figure \ref{fig:pic3}. Canonical Correlation Analysis (CCA) is a representative technique \cite{hotelling1936relations} that can find two linear projections making the multi-view data maximally correlated. We then obtain embeddings using the basis vectors of the two projections. Many techniques extend CCA to capture nonlinear inter-view relationships \cite{akaho2006kernel, andrew2013deep}. These methods, however, only support two-view data. Multi-view representation fusion can exploit the complementary information of multiple views to generate the required embeddings \cite{blei2003modeling, chen2010predictive, srivastava2012multimodal}. The principle is to determine the posterior probability p(h|x,y) of the probabilistic model p(x,y,z) over the joint space of the shared latent variables z and the observed two-view data x,y. Applying the neural network in multi-view representation learning is a recent trend. Representative examples are multi-modal autoencoder \cite{ngiam2011multimodal}, multi-view convolutional neural network \cite{feichtenhofer2016convolutional}, and multi-modal recurrent neural network \cite{karpathy2015deep}. Literature surveys \cite{li2018survey, Zhao2017Multi, Tang2017survey} include most recent works. Existing techniques require objects to have the same number of views. They thus cannot be used for arbitrary subsets with different numbers of features.


\subsection{Multi-dimensional Data Exploration}
Many visualization techniques aim at finding subsets where patterns exist \cite{shneiderman1996eyes}. A common strategy is projecting the data into a plane \cite{jackle2015temporal, lehmann2016general, bach2015time, elmqvist2008rolling}. The projection works as an overview, in which analysts manipulate data and filter parts of interest. Many works combine visualization and automatic algorithms to form a generic tool \cite{fujiwara2020visual, wenskovitch2017towards, sacha2016visual, loorak2015timespan, gleicher2013explainers}. These methods, however, have done good jobs in searching for subsets where patterns exist, but they cannot find patterns among a group of subsets by considering their feature similarity.

Subspace analysis is a kind of technique to find patterns in dimensional subspaces to overcome the problem of the Curse of Dimensionality \cite{beyer1999nearest}. A common strategy is to design a measure that describes the possibility that patterns exist within a subspace. Ferdosi \cite{ferdosi2011visualizing} proposes a measure for reordering axes of parallel coordinates to identify high-dimensional structures. Tukey et al. \cite{tukey1988computer} propose Scagnostics to identify anomalous scatterplots in the scatterplot matrix (SPLOM). The idea is to reduce the original SPLOM with $O(n^{2})$ cells (n is the number of attributes) to a scagnostics SPLOM with $O(k^{2})$ cells, where k is the number of measures that describe the distribution of points of an original SPLOM cell. Seo et al. \cite{seo2004rank} propose a rank-by-feature framework, in which users sort views according to a ranking measure. There are many other measures to find views of interest \cite{tatu2009combining, albuquerque2010improving, wongsuphasawat2015voyager}. They, however, are for evaluating views with 1D or 2D axes, i.e. low-dimensional subspaces.

Many approaches assist in exploring patterns in high-dimensional subspaces. Zhou et al. \cite{zhou2016dimension} propose a method to preserve clusters by reconstructing dimensions of subspaces. Wang and Mueller \cite{wang2017subspace} decompose a high-dimensional space into a series of 3D subspaces to facilitate pattern exploration. Yuan \cite{yuan2013dimension} proposes Dimension Projection Matrix/Tree that enables to explore record- and dimension-related patterns at the same time. Pagliosa et al. \cite{pagliosa2015projection} design an interactive tool for comparing different multi-dimensional projections. Xia et al. \cite{xia2017ldsscanner} design the LTSD-GD view that can represent latent low-dimensional structures within multi-dimensional data. Many methods design measures to evaluate how much insights are provided by a multi-dimensional projection \cite{motta2015graph, lehmann2015optimal, anand2012visual}. All these methods, however, work at the record level without the ability to find patterns associated with subsets.

Shadoan and Weaver \cite{shadoan2013visual} propose a visualization system for analyzing relationships between subsets. The relationship, however, reflects the common records between subsets, not as general as SEN encoding arbitrary subset features. Gratzl et al. \cite{gratzl2014domino} propose Domino that supports the flexible exploration of subsets and their relationships. Borland et al. \cite{borland2019selection} propose a visual analytics method to unbiasedly selecting a representative subset for a large dataset. Gotz et al. \cite{gotz2019visual} propose a method for interactively determining the most informative event subset in a specific analysis context. These methods, however, are different from our approach that aims at providing a generic tool to explore patterns among a large number of subsets. 


\section{Conclusion and Future Work}
\label{sec:section8} 

We have presented an approach to exploring multi-dimensional data at the subset level. The core of the approach is a subset embedding network that has three characteristics compared with existing embedding techniques. First, it supports arbitrary subsets with different numbers of features. Second, it captures the similarity of subsets on single features. Third, it has high efficiency by using a simple structure with few parameters. The network has been integrated into a visualization system that integrates three components to achieve a flexible and efficient workflow. We present example usage scenarios with real-world data and conduct multiple quantitative experiments to demonstrate the general applicability and effectiveness of our approach. 

In the future, we plan to make two improvements. First, we will apply the approach in more fields to thoroughly test its applicability. Second, we will further enrich the functions of the visualization system. For example, we would like to support customizing subset features or integrate more intelligent and automatic techniques to assist in selecting subsets in the projection.

\section*{Acknowledgements}

This work is supported by National Key R$\&$D Program of China (2018YFC0831700), NSFC project (61972278), Natural Science Foundation of Tianjin (20JCQNJC01620), and the Browser Project (CEIEC-2020-ZM02-0132). This work is also supported by Shanghai Municipal Science and Technology Major Project (2021SHZDZX0103, 2018SHZDZX01) and ZJLab.

\newcommand{\etalchar}[1]{$^{#1}$}

\bibliographystyle{eg-alpha-doi}


\newpage

\end{document}